\documentclass[11pt,twoside,a4paper]{article}

\newcommand{\reporttitle}{Algorithmic Trading Strategy Development and Optimisation}
\newcommand{\gid}{25}
\newcommand{\reportauthorOne}{OWEN NYO WEI YUAN}
\newcommand{\cidOne}{2301840}
\newcommand{\reportauthorTwo}{VICTOR TAN JIA XUAN}
\newcommand{\cidTwo}{2301787}
\newcommand{\reportauthorThree}{ONG JUN YAO FABIAN}
\newcommand{\cidThree}{2301898}
\newcommand{\reportauthorFour}{RYAN TAN JUN WEI}
\newcommand{\cidFour}{2301826}

\usepackage{ifxetex}
\usepackage{textpos}
\usepackage[numbers]{natbib}
\usepackage{kpfonts}
\usepackage[letterpaper,hmargin=2.cm,vmargin=2.0cm,includeheadfoot]{geometry}
\usepackage{ifxetex}
\usepackage{stackengine}
\usepackage{tabularx,longtable,multirow,subfigure,caption}%hangcaption
\usepackage{fancyhdr}
\usepackage{color}
\usepackage[tight,ugly]{units}
\usepackage{url}
\usepackage{float}
\usepackage[english]{babel}
\usepackage{amsmath}
\usepackage{graphicx}
\usepackage[colorinlistoftodos]{todonotes}
\usepackage{dsfont}
\usepackage{epstopdf} % automatically replace .eps with .pdf in graphics
\usepackage{natbib}
\usepackage{backref}
\usepackage{array}
\usepackage{latexsym}
\usepackage{etoolbox}

\usepackage{enumerate} % for numbering with [a)] format 

\usepackage{tabularx}

\ifxetex
\usepackage{fontspec}
\else
\usepackage[pdftex,pagebackref,hypertexnames=false,colorlinks]{hyperref} % provide links in pdf
\hypersetup{pdftitle={},
  pdfsubject={}, 
  pdfauthor={\reportauthorOne},
  pdfkeywords={}, 
  pdfstartview=FitH,
  pdfpagemode={UseOutlines},% None, FullScreen, UseOutlines
  bookmarksnumbered=true, bookmarksopen=true, colorlinks,
    citecolor=black,%
    filecolor=black,%
    linkcolor=black,%
    urlcolor=black}
\usepackage[all]{hypcap}
\fi

\usepackage{tcolorbox}

\usepackage{ntheorem}
\theoremstyle{break}

\ifxetex
\else
\fi

\def\@makechapterhead#1{%
  \vspace*{10\p@}%
  {\parindent \z@ \raggedright %\sffamily
    \interlinepenalty\@M
    \Huge \bfseries 
    \thechapter \space\space #1\par\nobreak
    \vskip 30\p@
  }}

\def\@makeschapterhead#1{%
  \vspace*{10\p@}%
  {\parindent \z@ \raggedright
    \sffamily
    \interlinepenalty\@M
    \Huge \bfseries  
    #1\par\nobreak
    \vskip 30\p@
  }}

\def\Beginboxit
   {\par
    \vbox\bgroup
	   \hrule
	   \hbox\bgroup
		  \vrule \kern1.2pt %
		  \vbox\bgroup\kern1.2pt
   }

\def\Endboxit{%
			      \kern1.2pt
		       \egroup
		  \kern1.2pt\vrule
		\egroup
	   \hrule
	 \egroup
   }

\newenvironment{boxit*}{\Beginboxit\hbox to\hsize{}}{\Endboxit}

\allowdisplaybreaks

\makeatletter
\newcounter{elimination@steps}
\newcolumntype{R}[1]{>{\raggedleft\arraybackslash$}p{#1}<{$}}
\def\elimination@num@rights{}
\def\elimination@num@variables{}
\def\elimination@col@width{}

\newcommand{\eliminationstep}[2]
{
    \ifnum\value{elimination@steps}>0\leadsto\quad\fi
    \left[
        \ifnum\elimination@num@rights>0
            \begin{array}
            {@{}*{\elimination@num@variables}{R{\elimination@col@width}}
            |@{}*{\elimination@num@rights}{R{\elimination@col@width}}}
        \else
            \begin{array}
            {@{}*{\elimination@num@variables}{R{\elimination@col@width}}}
        \fi
            #1
        \end{array}
    \right]
    & 
    \begin{array}{l}
        #2
    \end{array}
    &%                                    moved second & here
    \addtocounter{elimination@steps}{1}
}
\makeatother

\makeatletter  
\def\colvec#1{\expandafter\colvec@i#1,,,,,,,,,\@nil}
\def\colvec@i#1,#2,#3,#4,#5,#6,#7,#8,#9\@nil{% 
  \ifx$#2$ \begin{bmatrix}#1\end{bmatrix} \else
    \ifx$#3$ \begin{bmatrix}#1\\#2\end{bmatrix} \else
      \ifx$#4$ \begin{bmatrix}#1\\#2\\#3\end{bmatrix}\else
        \ifx$#5$ \begin{bmatrix}#1\\#2\\#3\\#4\end{bmatrix}\else
          \ifx$#6$ \begin{bmatrix}#1\\#2\\#3\\#4\\#5\end{bmatrix}\else
            \ifx$#7$ \begin{bmatrix}#1\\#2\\#3\\#4\\#5\\#6\end{bmatrix}\else
              \ifx$#8$ \begin{bmatrix}#1\\#2\\#3\\#4\\#5\\#6\\#7\end{bmatrix}\else
                 \PackageError{Column Vector}{The vector you tried to write is too big, use bmatrix instead}{Try using the bmatrix environment}
              \fi
            \fi
          \fi
        \fi
      \fi
    \fi
  \fi 
}  
\makeatother

\robustify{\colvec}

\ifxetex
\else
\fi
\ifxetex
\else

\fi
\definecolor{darkgreen}{rgb}{0,0.6,0}

\newcommand{\blue}[1]{{\color{blue}#1}}

\renewcommand{\emph}[1]{\blue{\bf{#1}}}

\begin{document}
% front page
% Last modification: 2016-09-29 (Marc Deisenroth)
% Modification for UW: 2017-05-22 (jphickey)
\begin{titlepage}

\newcommand{\HRule}{\rule{\linewidth}{0.5mm}} % Defines a new command for the horizontal lines, change thickness here

%----------------------------------------------------------------------------------------
%	LOGO SECTION
%----------------------------------------------------------------------------------------

\begin{center} % Center remainder of the page

%----------------------------------------------------------------------------------------
%	HEADING SECTIONS
%----------------------------------------------------------------------------------------

\includegraphics[width = 10cm]{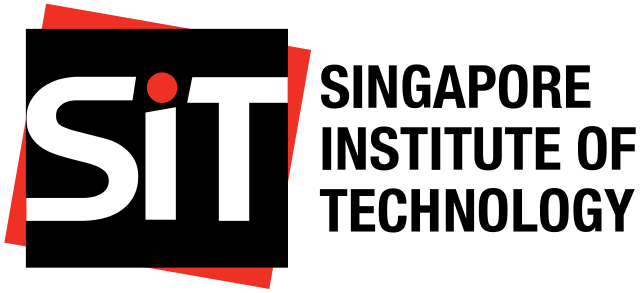}\\[1.5cm] 
\textbf{\textsc{\Large INF2006 - Cloud Computing and Big Data}}\\[1.0cm]
\textsc{\Large Singapore Institute of Technology}\\[0.5cm] 
\textsc{\large Infocomm Technology}\\[0.95cm] 

%----------------------------------------------------------------------------------------
%	TITLE SECTION
%----------------------------------------------------------------------------------------

\HRule \\[0.4cm]
{ \huge \bfseries \reporttitle}\\ % Title of your document
\HRule \\[1.5cm]
\end{center}

%----------------------------------------------------------------------------------------
%	AUTHOR + INSTRUCTOR SECTION (side by side)
%----------------------------------------------------------------------------------------
\begin{minipage}[t]{0.5\textwidth}
    \begin{flushleft} \large
        \textbf{Group \gid}\\
        \reportauthorOne~(ID: \cidOne)\\
        \reportauthorTwo~(ID: \cidTwo)\\
        \reportauthorThree~(ID: \cidThree)\\
        \reportauthorFour~(ID: \cidFour)\\
    \end{flushleft}
\end{minipage}
~
\begin{minipage}[t]{0.5\textwidth}
    \begin{flushright} \large
        \textbf{Instructors}\\
        Alvin Chan, \textit{\small Module Lead}\\
        Junhua Liu, \textit{\small Lead, Big Data}\\
        Kwang Yong Lim, \textit{\small Lab Instructor}\\
        Sheng Chen, \textit{\small Lab Instructor}\\
        Gopika Gopikrishnan, \textit{\small Teaching Assistant}\\
    \end{flushright}
\end{minipage}

\vspace*{\fill}
\begin{center}
    \makeatletter
    Date: \@date
    \makeatother
\end{center}

\end{titlepage}

%%%%%%%%%%%%%%%%%%%%%%%%%%% table of content
%If a table of content is needed, simply uncomment the following lines
\tableofcontents
\newpage

%%%%%%%%%%%%%%%%%%%%%%%%%%%% Main document
\section{Introduction}
\subsection{Project Overview}
    This project focuses on the development and optimisation of an enhanced algorithmic trading strategy crafted using historical S\&P 500 daily market data from the period 2000 through 2024 and quarterly earnings call transcripts. 
    
    \vspace{0.6em}
    
    The trading assignment notebook has provided a working baseline strategy built upon a 50 - day moving average indicator combined with FinBert sentiment classification. The overall objective of this assignment is to improve the baseline strategy with the enhanced strategy with the aid of additional data visualizations, computational optimisation, and a variety of back-testing procedures. 
    
    \vspace{0.6em}
    
    The development process starts with the provided dataset split: Development Set (2000 - 2017), Validation Set (2018 - 2024), and a fully concealed Test Set. The process then involves tweaking strategy, feature engineering, and fine-tuning multiple parameters within the development set to avoid overfitting to the validation set. When the results with the development set are satisfactory, the strategy will be performed on the validation set once to observe for generalisation. The fully held-out test set will serve as the final benchmark for a performance assessment. 
    
    \vspace{0.6em}
    
    The following metrics will be used for evaluation: Total Return, Sharpe Ratio, Maximum Drawdown, Win Rate, and Volatility to ensure a fair yet comprehensive assessment of risk and returns. In addition to financial performance, the project will also place an emphasis on computational scalability and efficiency, given the substantial volume of approximately two million rows of data in the development period. This ensures that the implementation must not only demonstrate well in the investment sector, but also in the execution and design sector.

\subsection{Motivation and Background}
    Algorithmic trading has emerged as a predominant force in contemporary financial markets, significantly influencing equity trading volume in developed markets [1]. The proliferation of high-frequency data and advancements in computational infrastructure have enabled systematic trading strategies to exploit statistical inefficiencies at scale. In contrast, traditional technical strategies rely primarily on historical price information and may exhibit lag effects, limiting their responsiveness to forward-looking market developments.
        
    \vspace{0.6em}
    
    Empirical finance research demonstrates that textual information, such as earnings call transcripts and managerial tone, contains predictive signals regarding future stock returns and volatility [2][3]. The rapid advancement of Natural Language Processing (NLP), particularly transformer-based architectures such as BERT [4], has facilitated structured sentiment extraction from unstructured financial documents. FinBERT, a domain-specific model fine-tuned for financial sentiment classification, has shown effectiveness in capturing earnings call tone and its influence on market reactions [5].
        
    \vspace{0.6em}
    
    Technical indicators serve to characterise price momentum, trend dynamics, and volatility regimes, while sentiment analysis provides forward-looking insights derived from managerial communication. Integrating quantitative technical indicators with qualitative sentiment measures enables the construction of a multi-factor investment framework. The combination of complementary information sources has the potential to improve signal reliability and reduce drawdowns relative to single-factor approaches.
        
    \vspace{0.6em}
    
    Furthermore, this project emphasises efficient big data processing methodologies. The dataset comprises millions of observations across numerous tickers spanning over two decades, rendering naive iterative implementations computationally inefficient. Consequently, vectorisation, memory optimisation, and scalable data processing pipelines are essential design considerations in both academic research and practical quantitative finance systems.
    
\subsection{Objectives}
    The primary objective of this project is to design and implement an enhanced algorithmic trading strategy that demonstrably outperforms the provided baseline across both financial performance and computational efficiency dimensions. Specifically, the project aims to:

    \begin{itemize}
        \item \textbf{Improve Risk-Adjusted Performance:} 
        Achieve a higher Sharpe Ratio than the baseline strategy, ensuring that any increase in return is accompanied by proportionate risk control rather than increased volatility exposure.
        
        \item \textbf{Enhance Downside Protection:} 
        Reduce Maximum Drawdown through refined entry, exit, and stop-loss mechanisms, thereby improving capital preservation during adverse market conditions.
        
        \item \textbf{Increase Signal Reliability:} 
        Strengthen trade signal precision by incorporating multiple technical indicators (e.g., momentum, volatility, and trend-based measures) beyond the baseline’s single 50-day moving average, and by requiring multi-signal confirmation before position entry.
        
        \item \textbf{Leverage Earnings Call Sentiment More Effectively:} 
        Improve the utilisation of FinBERT outputs by incorporating sentiment confidence scores, systematically handling neutral classifications, and evaluating multi-quarter sentiment consistency to reduce noise from isolated earnings events.
        
        \item \textbf{Refine Risk and Capital Allocation Rules:} 
        Develop improved position sizing logic and portfolio allocation mechanisms to manage concentration risk and enhance overall portfolio stability.
        
        \item \textbf{Ensure Generalization Across Time Periods:} 
        Demonstrate consistent performance improvements over the baseline across the Development, Validation, and Test datasets without evidence of overfitting.
        
        \item \textbf{Optimize Computational Efficiency and Scalability:} 
        Implement vectorised and memory-efficient computations capable of processing large-scale financial datasets efficiently, ensuring scalability beyond the current dataset size.
    \end{itemize}
    
    These goals are aimed at creating a strategy that not only outperforms the baseline in terms of performance metrics but is also solid in its methods, applicable in various situations, and reliable in computation.

\clearpage
\section{Enhanced Exploratory Data Analysis (EDA)}

    To gain deeper insights into the dataset beyond the baseline exploratory analysis, several additional analyses were conducted to examine the relationships between technical indicators and future stock returns. The analyses focused on momentum, trend, and volatility characteristics to determine whether these indicators exhibit consistent patterns associated with future 21-day returns, which could inform the design of the trading algorithm.

\subsection{Momentum vs Future Returns}

    A 63-day momentum indicator was calculated for each stock using the percentage change in closing price over the previous 63 trading days. This indicator was compared with the forward 21-day return to evaluate the predictive relationship between momentum and future performance (see Figure~\ref{fig:momentum_vs_returns} in Appendix A).
    
    \vspace{0.6em}
    
    The calculated correlation of approximately 0.019 indicates that the relationship between momentum and future returns is weak. This suggests that momentum alone may not be an effective predictive variable for modelling purposes. However, momentum may still be useful as a signal for identifying stocks with potential to outperform.

\subsection{Cumulative Returns}

    The cumulative return comparison between the top momentum stocks and the overall market average is illustrated in Figure~\ref{fig:cumulative_returns} (Appendix A). It shows that portfolios consisting of high-momentum stocks generally outperform the broader universe over time.

\subsection{200-Day Moving Average}

    Stocks were categorised into two groups based on whether their prices were above or below the 200-day moving average (MA200). The average forward 21-day returns were then calculated for both categories.
    
    \vspace{0.6em}
    
    Average 1-month forward return:
    \begin{itemize}
        \item Above MA200: 0.012556582603665217
        \item Below MA200: 0.01724687203601836
    \end{itemize}

    Stocks below MA200 produce slightly higher forward returns than those above it. This could suggest a short-term rebound trend effect (revision). This suggests that stocks below the long-term trend may experience short-term rebound effects, while stocks above MA200 continue to follow longer-term upward trends.
    
\subsection{Average True Range (ATR)}

    Volatility was examined using the Average True Range (ATR), calculated over a 14-day exponential moving average of the true range. ATR measures the typical daily price movement and serves as a proxy for market volatility. The relationship between ATR volatility and forward returns is visualized in Figure~\ref{fig:atr_volatility} (Appendix A).
    
    \vspace{0.6em}
    
    The analysis shows a high concentration of observations at lower ATR values, indicating that most stocks in the dataset experience relatively low volatility. As ATR increases, the number of observations decreases and the spread of forward returns becomes wider. This suggests that higher volatility stocks exhibit greater variability in future returns.
    
    \vspace{0.6em}
    
    From the scatter plot, it can also be observed that most returns cluster around zero regardless of volatility level. However, stocks with higher ATR values tend to show more extreme positive and negative forward returns. This indicates that while higher volatility does not necessarily guarantee higher returns, it is associated with greater uncertainty and larger potential price movements.
    
\subsection{Momentum Decile Analysis}

    Decile analysis was conducted by ranking stocks according to their 63-day momentum and comparing these groups with the next 21-day returns. The results of this analysis are shown in Figure~\ref{fig:momentum_decile} (Appendix A).
    
    \vspace{0.6em}
    
    The results show that stocks in the lowest momentum decile (Decile 1) and highest momentum decile (Decile 10) generate the highest subsequent returns. Stocks in the highest momentum decile tend to continue outperforming due to momentum persistence, while stocks in the lowest momentum decile often experience short-term rebounds after periods of extreme underperformance.

\subsection{ATR Bucket Analysis}
    
    The analysis of ATR percentage buckets versus the next 21-day returns shows a clear upward trend. Stocks in the highest ATR bucket (Decile 10) generate the largest forward returns, while those in the lowest volatility bucket (Decile 1) produce the lowest forward returns. The average forward returns across ATR volatility buckets are visualized within Figure~\ref{fig:atr_bucket} (Appendix A).
    
    \vspace{0.6em}
    
    The trend is observed in both the development and validation datasets, where higher ATR buckets consistently show increasing average forward returns. This suggests that stocks experiencing larger price movements may offer greater short-term return opportunities, although they are also associated with higher risk.

\subsection{Momentum-to-Volatility Ratio}
    
    Further analysis was conducted using the momentum-to-volatility ratio (momentum divided by ATR). The decile analysis of this ratio indicates that stocks in the highest bucket (Decile 10) exhibit stronger forward returns. The results can be visualized in Figure~\ref{fig:mom_atr_ratio} (Appendix A).
    
    \vspace{0.6em}
    
    This suggests two important observations. First, stocks with high momentum relative to their volatility tend to generate stronger short-term returns. Second, combining momentum with volatility may help identify stocks that exhibit strong trends while maintaining more stable price movements, making them potentially more reliable trading opportunities.

\section{Data Cleaning and Preparation}

    Before performing feature engineering and strategy development, both the stock price dataset and the earnings transcript dataset were examined for potential data quality issues. Several preprocessing steps were applied to ensure that the data was reliable and consistent.
    
\subsection{Duplicate Removal and Date Standardisation}

    Duplicate records were removed from both datasets. Date fields were converted into a standardised datetime format to ensure that entries could be ordered correctly for time-series analysis.

\subsection{Handling Missing Price Values}
    Missing price values were handled using forward filling within each ticker group to maintain a continuous time series. This approach replaces missing values with the most recent available observation for the same stock, preserving realistic price trends without introducing artificial values. After forward filling, any remaining rows containing missing price values were removed.

\subsection{Filtering Tickers with Insufficient Historical Data}

    Some tickers contained very limited trading histories, which could lead to unreliable calculations for technical indicators such as long-term moving averages, momentum indicators, and volatility measures.

    \vspace{0.6em}
    
    To address this, a minimum data requirement of three years of trading history (approximately 756 trading days) was imposed. Tickers that did not meet this threshold were removed from the dataset.

\subsection{Outlier Detection and Removal}

    Financial datasets may contain extreme or erroneous price movements that could distort statistical analysis. Daily returns were calculated using percentage changes in closing prices for each ticker.

    \vspace{0.6em}
    
    Observations with extremely large daily price movements exceeding 80\% were considered unrealistic and were removed from the dataset.

\subsection{Cleaning Earnings Transcript Data}

    The earnings transcript dataset contains textual data used for sentiment analysis. Records with missing transcripts or transcripts shorter than 100 characters were removed, as they likely contained insufficient information for meaningful sentiment evaluation.
    
    \vspace{0.6em}
    
    To maintain consistency between datasets, the earnings data were further filtered to include only tickers that remained in the cleaned price dataset. This ensures alignment between price data and earnings sentiment data for later analysis.
    
    \vspace{0.6em}
    
    These data preparation steps improve overall data quality and ensure that subsequent analysis is performed on a reliable and well-structured dataset.

\section{Computational Efficiency and Performance Optimisation}
\subsection{Scalability Challenges}
\subsubsection{Large Time-Series Dataset}
    As the dataset comprises over two million price records across hundreds of tickers spanning more than two decades, some of the challenges include rolling indicators such as MA200, EMA50, ATR14, and Momentum63, which require historical windows for each ticker. Additionally, calculations must be performed repeatedly across all trading days, and time-series operations are computationally expensive when implemented in loops. Without proper optimisation in place, it can lead to long runtime, high memory consumption and inefficient processing of multi-ticker datasets.

\subsubsection{Cross-Sectional Ranking Computation}
    As the strategy performs a weekly ranking of all stocks based on the 63-day momentum indicator, there are two key scalability challenges. Firstly, the ranking must be computed for every evaluation date. Secondly, each ranking operation requires all the tickers to be processed simultaneously, increasing computational overhead as the dataset grows.

\subsubsection{Indicator Recalculation During Backtesting}
    During backtesting, the trading algorithm repeatedly evaluates entry and exit conditions for each trading day. If technical indicators such as moving averages, momentum, and ATR are recalculated during each evaluation step, this can lead to redundant computations and significantly slow down the simulation process.

\subsection{Optimisation Strategies}
\subsubsection{Vectorised Indicator Computation}
    To improve computational efficiency, technical indicators such as MA200, EMA50, EMA200, Momentum63 and ATR14 were implemented using vectorised operations in Pandas. Functions such as groupby().transform(), rolling(), and ewm() allow calculations to be executed at the column level rather than iterating through rows with Python loops. This approach significantly reduces runtime as vectorised operations are executed in optimised compiled code.

\subsubsection{Pre-Computation of Technical Indicators}
    All the technical indicators were calculated during the data preprocessing stage rather than being recomputed during the trading simulation. By precomputing indicators such as moving averages, volatility measures, and momentum values, the trading algorithm only needs to evaluate entry and exit conditions during backtesting. This eliminates redundant computations and improves the overall efficiency of the simulation process.

\subsubsection{Pre-Computation of Technical Indicators}
    Daily price data was compressed into weekly frequency for selected components of the strategy during ranking and signal evaluation stages to improve computational efficiency. Given that the strategy performs cross-sectional ranking and portfolio rebalancing on a weekly basis, operating on daily data introduced redundant computations without providing additional decision-making value. By aggregating the daily prices into weekly prices (end-of-week closing prices), the number of data rows needed to be processed is significantly reduce by six times.

    This optimisation provided the following several benefits as well:
    \begin{itemize}
        \item \textbf{Reduced Computational Load:} The dataset size decreases by approximately 80\%, leading to faster execution of ranking, filtering, and backtesting operations.
        
        \item \textbf{Faster Iteration Cycles:} Strategy experimentation and parameter tuning can be performed more efficiently due to shorter runtime.
        
        \item \textbf{Preserved Signal Integrity:} Since trading decisions are executed weekly, the use of weekly data does not materially affect strategy behaviour or performance.
    \end{itemize}

    To conclude the optimisation section, temporal data compression complements other optimisation techniques such as vectorisation and pre-computation, enabling the ability to process large amounts of financial dataset while retaining the trading strategy outcomes.
    
\subsection{Performance Measurements}
\subsubsection{Runtime Improvements}
    The implementation of vectorised operations and precomputed indicators significantly reduces the time required for data processing and backtesting. Compared to a naive loop-based implementation, the optimised pipeline completes indicator computation and strategy simulation within a few minutes, enabling multiple strategy experiments to be conducted efficiently.

\subsubsection{Backtesting Efficiency}
    The optimisation techniques allow the complete backtesting process across the full development dataset, ranging from 2000 to 2017, to be executed efficiently despite the large dataset size. The removal of redundant computations and caching of ranking results significantly reduces the computational load during the trading simulation.

\clearpage
\section{Strategy Enhancement Methodology}
\subsection{Technical Indicators and Features}
    The enhanced strategy extends beyond the baseline 50-day moving average by introducing a structured multi-layer signal framework designed to improve trend alignment, alpha concentration, and volatility-adjusted risk control.
    
    \vspace{0.6em}
    
\textbf{Long-Term Regime Filter (MA200)}

    A 200-day simple moving average (MA200) was implemented as a macro-regime filter. Long trades are considered only when:
    
    \[
    Price > MA200
    \]

    This ensures that positions are entered exclusively during structurally bullish environments and reduces exposure to prolonged bear markets. Compared to the baseline, which relies on a shorter-term average, MA200 introduces stronger macro-trend alignment.
    
    \vspace{0.6em}
    
\textbf{Trend Confirmation (EMA50 and EMA200)}

    Two exponential moving averages were introduced:
    
    \begin{itemize}
        \item EMA50 (fast trend indicator)
        \item EMA200 (slow trend indicator)
    \end{itemize}
    
    A bullish trend condition requires:
    
    \[
    EMA50 > EMA200 \quad \text{and} \quad Price > EMA50
    \]
    
    The exponential formulation increases responsiveness compared to simple moving averages while retaining trend stability. This dual confirmation prevents premature entries during weak rebounds and ensures alignment between intermediate and long-term trend dynamics.
        
    \vspace{0.6em}

\textbf{Momentum Filter (63-Day Return)}

    A 63-day momentum indicator (approximately three months) was incorporated:
    \[
    Momentum_{63} = \frac{Price_t}{Price_{t-63}} - 1
    \]
    Only stocks satisfying:
    \[
    Momentum_{63} > 0
    \]
    are eligible for entry.
            
    \vspace{0.6em}

    This filter ensures that trades are taken only in stocks demonstrating intermediate-term strength, reinforcing continuation effects and eliminating stagnant or declining assets.
            
    \vspace{0.6em}

\clearpage
\textbf{Cross-Sectional Relative Strength Ranking (Top-N Filter)}

    A key enhancement is the introduction of cross-sectional momentum ranking.
    For each weekly evaluation date:
    
    \vspace{0.6em}
    
    \begin{enumerate}
        \item The most recent available analytics row (as-of logic) is selected per ticker.
        \item Tickers are ranked by descending 63-day momentum.
        \item Only the Top $N$ stocks ($N = 10$) are eligible for entry.
    \end{enumerate}
    
    This mechanism concentrates capital into the strongest momentum leaders and prevents diversification across average-performing stocks. By limiting eligibility to the Top 20 names, the strategy deliberately emphasises alpha concentration.
    To ensure computational efficiency, the Top-N results are cached per evaluation date, preventing redundant ranking calculations during weekly decision cycles.
            
    \vspace{0.6em}

\textbf{Volatility-Based Risk Control (ATR14 Trailing Stop)}

    The Average True Range (ATR14) was implemented as a dynamic trailing stop mechanism.
    True range is defined as:
    \[
    TR = \max(high - low,\ |high - prev\_close|,\ |low - prev\_close|)
    \]
    ATR14 is computed using exponential smoothing. A trailing stop is defined as:
    \[
    Stop = HighestCloseSinceEntry - (3.5 \times ATR_{14})
    \]
    If the price falls below this level, the position is exited.
    Unlike the baseline’s fixed 20\% stop-loss, this volatility-adjusted mechanism adapts to changing market conditions, tightening during low volatility and widening during high volatility regimes.
                
    \vspace{0.6em}

\subsection{Earnings Transcript Analysis}
\textbf{FinBERT Sentiment Extraction}
    Earnings transcripts are processed using a FinBERT pipeline. For efficiency and relevance:

    \begin{itemize}
        \item Only the final 2000 characters of each transcript are analysed, prioritising recent managerial commentary.
        \item The model returns a categorical sentiment classification: positive, neutral, or negative.
        \item If transcript data or the pipeline is unavailable, sentiment is ignored.
    \end{itemize}
                
    \vspace{0.6em}
    
\textbf{Sentiment as a Gating Mechanism}

    Sentiment is not used as a direct entry trigger but as a risk filter. If earnings sentiment is classified as negative, all BUY signals for that ticker are blocked regardless of technical strength. Neutral sentiment does not prevent entry. This design ensures that strong technical setups are not overridden by adverse fundamental signals while avoiding excessive sensitivity to minor sentiment variations.

\clearpage
\subsection{Entry, Exit, and Risk Management Rules:}
\textbf{Entry Conditions (BUY)}

    A BUY signal is generated only when all of the following conditions are satisfied:

    \begin{enumerate}
        \item $Price > MA200$ (bullish regime filter)
        \item $EMA50 > EMA200$ and $Price > EMA50$ (trend confirmation)
        \item $Momentum_{63} > 0$
        \item Ticker is ranked within the Top 10 by 63-day momentum
        \item Earnings sentiment is not negative
    \end{enumerate}

    This multi-layer confirmation structure reduces false positives by requiring alignment across macro regime, intermediate trend, momentum strength, cross-sectional leadership, and fundamental sentiment.
            
    \vspace{0.6em}

\textbf{Exit Conditions (SELL)}

    A position is exited when any of the following conditions are met:
    
    \begin{enumerate}
        \item $Price < MA200$ (regime breakdown)
        \item $Price < HighestCloseSinceEntry - (3.5 \times ATR14)$ (volatility trailing stop)
    \end{enumerate}
    
    The layered exit logic provides both structural protection (trend breaks) and adaptive volatility-based protection (ATR stop), improving drawdown control relative to a fixed-percentage stop.
            
    \vspace{0.6em}

\textbf{Computational Design Considerations}

    Given the scale of the data set (approximately two million price records), efficiency considerations were embedded in the implementation:
    
    \begin{itemize}
        \item Rolling and exponential indicators are computed using vectorized \texttt{groupby().transform()} operations.
        \item Exponential weighted moving averages (\texttt{ewm}) avoid iterative loops.
        \item Cross-sectional Top-N ranking is cached per evaluation date to reduce redundant computation.
        \item As-of ranking logic prevents look-ahead bias while maintaining efficiency.
    \end{itemize}
    
    These optimisations ensure the strategy remains scalable and suitable for large financial datasets without significant runtime degradation.

\section{Experimental Results and Analysis}

\subsection{Development Phase Experiments}

\subsubsection{Momentum Predictive Power}

\textbf{Hypothesis}

Stocks exhibiting strong recent momentum are more likely to continue performing well in the near future. Specifically, stocks with higher 63-day price momentum are expected to generate higher forward returns over the next month.
\vspace{0.6em}

\textbf{Methodology}

Momentum was calculated as the percentage change in closing price over the previous 63 trading days. To evaluate the predictive power of momentum, the forward return over the next 21 trading days was computed. The team generated a scatterplot to visualise the relationship between the 63-day momentum and the subsequent 21-day forward return. Additionally, the correlation coefficient between the two variables was calculated to quantify the strength of the relationship.
\vspace{0.6em}

\textbf{Results}

The analysis revealed a positive relationship between the 63-day momentum and the forward returns, indicating that stocks with stronger recent performance tend to continue outperforming in the short term. Although the correlation was moderate, the overall trend supported the momentum hypothesis.
\vspace{0.6em}

\textbf{Implication for Strategy}

Based on these findings, the 63-day momentum was incorporated as a core signal in the enhanced trading strategy, where stocks with higher momentum values are prioritised when selecting potential investment candidates. 

\subsubsection{Long-Term Trend Filter Using Moving Averages}

\textbf{Hypothesis}

Stocks trading above their long-term moving averages are more likely to maintain positive performance trends compared to those trading below the moving average.
\vspace{0.6em}

\textbf{Methodology}

A 200-day moving average (MA200) was calculated for each stock. Stocks were then classified into 2 groups of prices above MA200 and those with prices below MA200. The average forward 21-day return was then calculated for both groups to determine whether the long-term trend indicator provided useful predictive information.
\vspace{0.6em}

\textbf{Results}

The results showed that stocks trading above the MA200 generally produced higher average forward returns, whilst stocks trading below the MA200 showed weaker or negative returns, suggesting that the long-term trend serves as a useful indicator of market regime and stock strength.
\vspace{0.6em}

\textbf{Implication for Strategy}

 A trend confirmation filter was incorporated into the enhanced strategy, where positions are only entered when the stock exhibits a strong upward trend, which is defined by 2 indicators: the EMA50 is greater than the EMA200, and the price is greater than the EMA50. This helps ensure that trades are executed during favourable market conditions.

 \subsubsection{Volatility Analysis using ATR}

\textbf{Hypothesis}

High volatility stocks may exhibit weaker or less predictable future returns
\vspace{0.6em}

\textbf{Methodology}

Volatility was measured using the Average True Range (ATR) over 14 days. ATR was normalised by dividing it by the stock price to obtain ATR\%, allowing comparison across stocks with different price levels. Stocks were then grouped into volatility buckets, and their forward 21-day returns were analysed to determine whether volatility had predictive power.
\vspace{0.6em}

\textbf{Results}

The results suggested that stocks with extremely high volatility did not consistently produce higher forward returns and were often associated with greater return dispersion, indicating that volatility may increase risk without necessarily improving expected returns.
\vspace{0.6em}

\textbf{Implication for Strategy}

 AR was incorporated into the strategy as a risk management tool rather than a selection signal. Specifically, an ATR-based trailing stop mechanism was implemented to limit downside risk, allowing profitable trades to continue while protecting the portfolio against large losses.

\clearpage
\section{Final Performance Outcomes}
    The final evaluation of the trading strategy was conducted using the datasets provided. The performance was assessed using five key metrics: Total Return, Sharpe Ratio, Maximum Drawdown, Win Rate, and Volatility, which collectively measure profitability, risk-adjusted performance, downside risk exposure, and trading consistency.

\subsection{Comprehensive Metrics}
    Table 1 below summarises the performance of the baseline strategy and the enhanced strategy across the development, validation, and test datasets.

\begin{table}[H]
\centering
\resizebox{\textwidth}{!}{
\begin{tabular}{|c|c|c|c|c|c|c|}
\hline
Dataset     & Strategy & Total Return & Sharpe Ratio & Maximum Drawdown & Win Rate & Volatility \\ \hline
Development & Baseline & 299.84\%     & 1.19         & -57.41\%         & 33.30\%  & 37.05\%    \\ \hline
Development & Enhanced & 466.14\%     & 1.69         & -23.08\%         & 46.60\%  & 30.28\%    \\ \hline
Validation  & Baseline & 12.45\%      & 0.41         & -35.42\%         & 27.75\%  & 43.80\%    \\ \hline
Validation  & Enhanced & 189.10\%     & 2.04         & -23.50\%         & 38.50\%  & 40.00\%    \\ \hline
Test        & Enhanced &   152.40\%   & 2.07     &  -13.5\%         &  41.0\%  &  26.10\%    \\ \hline
\end{tabular}
}
\caption{Performance Comparison of Baseline and Enhanced Strategies on All Datasets}
\end{table}

In the development dataset, the baseline strategy had an overall return of 299.84\%, a Sharpe Ratio of 1.19 and a maximum drawdown of -57.41\%. Although this shows that the base strategy could achieve a high level of returns in the long-term development period, the high level of drawdown shows a high level of exposure to downside risk in the unfavourable market conditions.

\vspace{0.6em}

Compared with the baseline, the improved strategy delivered a total return of 466.10\% on development data, representing a significant improvement. Moreover, the Sharpe Ratio has grown to 1.69, which represents a better risk-adjusted performance. The maximum drawdown was decreased to -23.10\%, indicating that the improved strategy had a significantly better impact on mitigating losses when the markets declined. The improved strategy also recorded a win rate of 0.47 and annual volatility of 30.3\%, which indicates a fairly balanced risk-reward profile.

\vspace{0.6em}

In the validation dataset, the baseline strategy had a total return of 12.45\%, a Sharpe Ratio of 0.41, a maximum drawdown of -35.42\%, a win rate of 0.28\%, and a volatility of 43.8\%. These findings imply that the baseline strategy was not generalising, which gave weaker risk-adjusted returns and greater volatility.

\vspace{0.6em}

The enhanced strategy has done much better in the validation dataset and earned a total return of 189.10\% and a Sharpe ratio of 2.04, and this means the strategy has performed well with regard to risk-adjusted performance. The maximum drawdown was kept at -23.50 percent with the win rate going to 0.39 and volatility being 40.0. These findings indicate that the improved strategy could generalise well on the basis of the development dataset without showing any overfitting.

\subsection{Comparative Analysis}
    Overall, the enhanced strategy outdid the baseline strategy in both datasets. The most notable improvements were observed in total return, Sharpe ratio, and drawdown control.
    
    \vspace{0.6em}
    
    Firstly, the enhanced strategy generated higher returns when compared to the baseline strategy. On the development dataset, total return increased from 299.84\% to 466.10\%, while on the validation dataset, the returns increased from 12.45\% to 189.10\%. This indicates that the additional signals and filtering mechanisms introduced in the enhanced strategy enhanced the model’s ability to detect profitable trading opportunities.
    
    \vspace{0.6em}
    
    Secondly, the enhanced strategy showed significantly improved risk-adjusted performance, as shown by the Sharpe ratio. The Sharpe ratio improved from 1.19 to 1.69 in the development dataset and from 0.41 to 2.04 in the validation dataset. This shows that the improved strategy not only provided higher returns, but it did so with better risk utilisation.
    
    \vspace{0.6em}
    
    Thirdly, the enhanced strategy significantly reduced the maximum drawdown. The baseline strategy showed a large drawdown of -57.41\% during the development phase, while the enhanced strategy reduced this to -23.10\%.  The same improvements were also recorded in the validation dataset. The reduced drawdowns suggest that the enhanced strategy implemented better effective risk management and avoided holding positions during prolonged market downturns.
    
    \vspace{0.6em}
    
    Many of these improvements are due to the multi-factor decision framework of the improved strategy, which incorporates the momentum indicators, volatility filtering based on the Average True Range (ATR) momentum to volatility ratio analysis and sentiment based on the earnings transcripts. The strategy ensures the reduction of false signals and the enhancement of the quality of the trade by having several indicators in place before any trades are made.

\subsection{Performance Visualization}
    To further analyse the behaviour of both strategies over time, several performance visualizations were generated.
    
    \begin{figure}[H]
        \centering
        \includegraphics[width=0.7\textwidth]{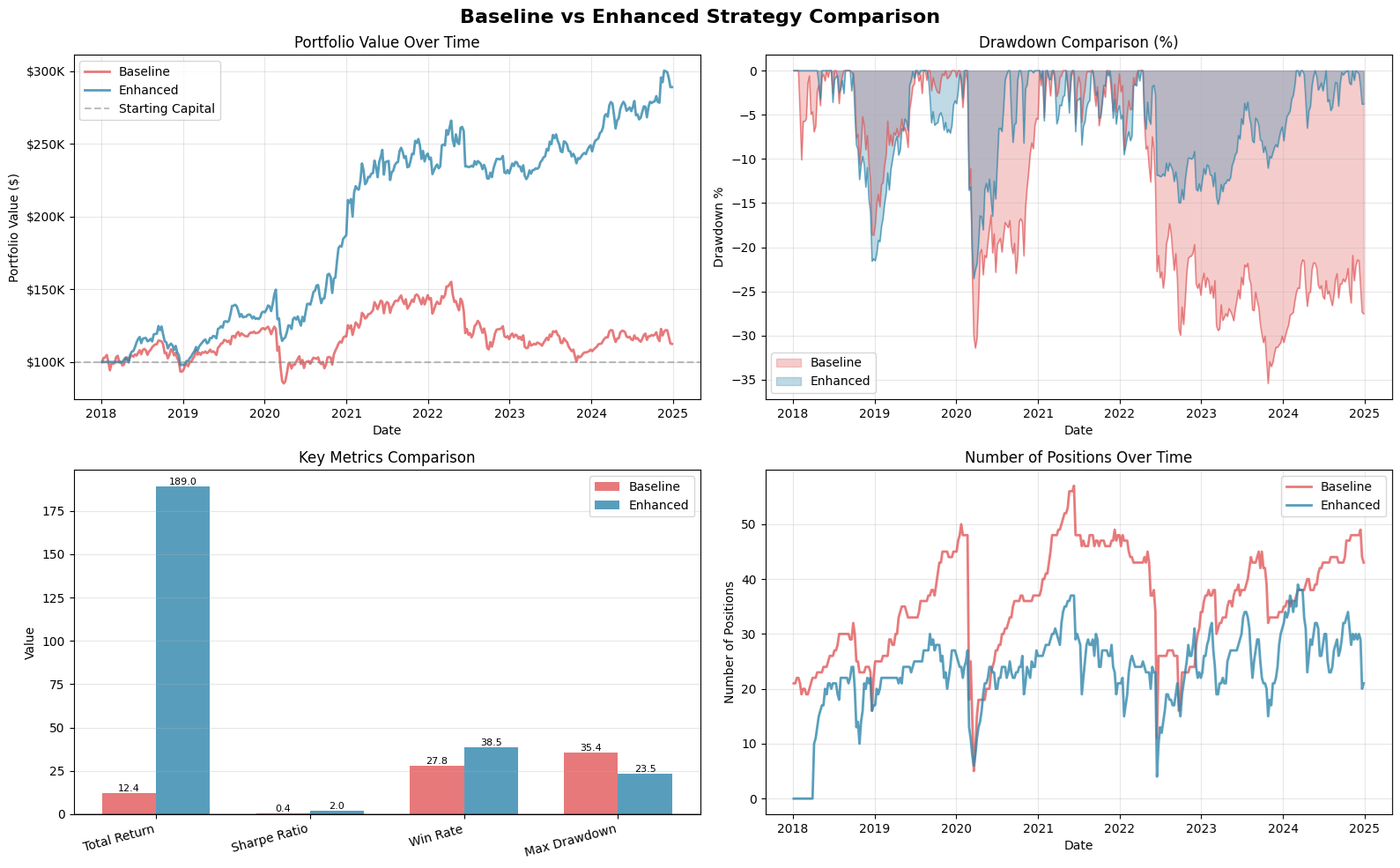}
        \caption{Baseline vs Enhanced Strategy Comparison}
        \label{fig:baseline-vs-enhanced-strategy-comparison}
        \end{figure}

\subsubsection{Portfolio Value over Time}

    The portfolio value chart is used to show the total value of the trading portfolio at each point in time during the simulation. Using an initial capital of \$100,000, the chart tracks how the portfolio evolves over time.

    \vspace{0.6em}

    In the case of the baseline strategy, the portfolio value shows an upward trend with several major declines, indicating fluctuating performance and sensitivity to market crashes. In contrast, the improved strategy exhibits a steadier upward trend, with smaller changes in portfolio value. This shows that the extra signals included in the improved strategy enhanced the capacity of the model to identify profitable trends and eliminate bad trading opportunities.

\subsubsection{Drawdown Comparison (\%)}

    The drawdown comparison chart shows the percentage change of the baseline strategy and the enhanced strategy of the previous peak value of the portfolio over time. According to the visualization, the baseline strategy has more in-depth and longer-term drawdowns, with the development dataset having the maximum drawdown of about -57.41\%. In contrast, the improved strategy has much lower drawdowns of approximately -23\%, which indicates a better preservation of the capital. This has been made possible by the multi-layer risk management framework of the improved strategy, trend filters and the ATR-based trailing stop, which helps exit positions earlier during the deteriorating market environments.

\subsubsection{Key Metrics Comparison}

    The key metrics comparison chart gives a graphical overview of the performance differences between the baseline and improved strategies with respect to multiple assessment measures, such as the total return, Sharpe ratio, win rate, and maximum drawdown.

    \vspace{0.6em}

    Based on the visualization, it is evident that the improved strategy performs better in comparison to the baseline strategy in most of the metrics. The enhanced strategy, in particular, generates overall greater returns and Sharpe ratios, which indicate increased profitability and improved risk-adjusted performance. The enhanced strategy also has a better win rate and significantly reduced maximum drawdown, thus showing that the multi-factor trading framework is effective in enhancing both the generation and the control of returns and risk.

\subsubsection{Number of Positions over Time}

    The number of positions over time chart shows the number of stocks actively held in the portfolio at any given time during the simulation period. This chart gives an understanding of portfolio diversification and the number of times that the strategy enters and exits. 

    \vspace{0.6em}
    
    The baseline strategy is more likely to have a bigger and more volatile number of positions over time and less selective conditions for entry. In contrast, the improved strategy has a more consistent and regulated number of positions because of the Top-N momentum ranking mechanism, restricting the trades to the most performing stocks. This narrow selection strategy enhances efficiency in capital allocation and helps to increase the overall performance of the improved strategy, and reduces drawdown.

\section{Conclusion and Future Work}
\subsection{Conclusion}
    This Project set out to develop an enhanced algorithmic trading strategy designed to improve upon the provided baseline model through the integration of additional technical indicators, sentiment analysis, and faster computational design. The original baseline approach revolved around a single 50-day moving average combined with the sentiment gate from FinBERT signals. While it provided decent results, the approach could be improved in many aspects, such as trend confirmation, cross-sectional selection, and volatility-adjusted risk management. Thus, the goal of the enhanced strategy was to address these drawbacks by providing a multi-layer signal approach.
            
    \vspace{0.6em}
        
    The first improvement involved stronger trend identification and market regime filtering. A 200-day moving average (MA200) was introduced to ensure trades were taken only during bullish market conditions. In addition, exponential moving averages (EMA50 and EMA200) were used to confirm trend direction, ensuring that trades align with both intermediate and long-term market trends.
            
    \vspace{0.6em}
        
    Momentum-based stock selection was also incorporated through a 63-day momentum indicator. Stocks were ranked using cross-sectional momentum, and only the Top-N candidates were selected for trading. This approach concentrates capital on the strongest performing assets rather than spreading it across weaker opportunities.
                
    \vspace{0.6em}
        
    Risk management was enhanced through an ATR-based trailing stop, which dynamically adjusts according to market volatility. Compared to the baseline’s fixed stop-loss, this approach provides better downside protection while reducing premature exits during volatile market conditions.
        
    \vspace{0.6em}
    
    To add on, the strategy incorporates sentiment extracted from earnings calls through the use of FinBERT. However, instead of acting as a direct signal for entry and exit, it is used more as a filter that blocks poor trades that align with negative managerial sentiment, while allowing trades with positive tone to proceed.
        
    \vspace{0.6em}
    
    Lastly, computational efficiency was a key priority as well, given the large amount of data that spans across two decades. By vectorising certain operations, reducing the amount of group-based calculations, and caching ranking procedures, the strategy was able to process the millions of records faster.
            
    \vspace{0.6em}
    
    Overall, the enhanced strategy was able to demonstrate that by combining trend filters, momentum selection, volatility-adjusted risk, and sentiment analysis, it was able to produce a more consistent framework compared to the baseline model. 

\subsection{Limitations}
    The first limitation lies in the number of technical indicators combined in the strategy. While the strategy has considered trend, momentum, and macro regime effectively, it does not account for volume-based signals, volatility regime, or market breadth indicators. With proper testing, these features could potentially provide richer insights into the market dynamics and improve the robustness of the signals produced by the trading algorithm.
    
    \vspace{0.6em}
        
    The second limitation relates to the simplified usage of sentiment information. The current implementation only uses a categorical sentiment label (positive, neutral, or negative) derived from the final portion of earnings transcripts. While this approach captures high-level managerial tone, it does not utilise sentiment intensity, contextual information, or topic-level analysis. As a result, some informative signals embedded within earnings discussions may not be fully captured by the model.
    
    \vspace{0.6em}
        
    Lastly, while vectorisation and caching techniques did improve computational efficiency, large-scale experimentation and parameter tuning remained computationally expensive. Running extensive simulations with multiple parameters across decades of historical data required significant processing time.
            
    \vspace{0.6em}
    
    \subsection{Challenges Encountered}
    Throughout the development phase, several technical and analytical challenges were encountered along the way. One major challenge involved ensuring that all calculations avoided look-ahead bias when computing indicators and ranking signals. Careful selection of 'as - of' logic ensured that only information available at the time of making was utilised within the strategy.
    
    \vspace{0.6em}
    
    Another challenge faced was the handling of large-scale financial datasets. Processing millions of rows of historical price data while consuming earnings transcript information meant that efficient data pipelines had to be utilised. In order to maintain acceptable runtime, the operations have to be vectorised, and nested loops have to be removed as well.
                
    \vspace{0.6em}
    
    In addition, sentiment coverage across earnings transcripts was not always consistent. Some tickers lacked transcript data for certain periods, requiring the strategy to gracefully handle missing sentiment information without introducing bias into the trading decisions.
    
    \vspace{0.6em}

\subsection{Future Work}
    Several promising directions can further improve the strategy’s effectiveness and research depth.
        
    \vspace{0.6em}
    
    Firstly, incorporating advanced NLP techniques for earnings transcript analysis. Instead of a basic sentiment classification, future work could include contextual embeddings, topic modelling, or sentence-level analysis to capture richer insights from earnings calls. This may reveal specific themes such as revenue guidance, cost pressures, or management confidence that could influence stock performance.
    
    \vspace{0.6em}
        
    Secondly, introducing machine learning models for signal combination could be introduced. Rather than manually defining entry and exit rules, supervised learning algorithms such as gradient boosting and random forest classification could be trained to learn about the relationships between technical indicators, sentiment signals, and future returns. This could potentially enhance the predictive power of the trading algorithm.
    
    \vspace{0.6em}
    
    Lastly, future works may look to explore on dynamic portfolio allocation methods. The strategy inherently did not allow for changes to be made towards the capital allocation, hence it allocated a fixed capital across the Top-N selection of momentum leaders. More sophisticated approach such as risk parity allocation, volatility - adjusted capital allocation, could help to optimise the portfolio and distribute capital more efficiently across selected assets.
    
    \vspace{0.6em}

\clearpage
\nocite{*}
\bibliography{mybib}

@article{kirilenko2017flash,
  title={The Flash Crash: High-Frequency Trading in an Electronic Market},
  author={Kirilenko, Andrei A. and Kyle, Albert S. and Samadi, Mehrdad and Tuzun, Tugkan},
  journal={The Journal of Finance},
  volume={72},
  number={3},
  pages={967--998},
  year={2017},
  publisher={Wiley},
  doi={10.1111/jofi.12498}
}

@article{loughran2011liability,
  title={When Is a Liability Not a Liability? Textual Analysis, Dictionaries, and 10-Ks},
  author={Loughran, Tim and McDonald, Bill},
  journal={The Journal of Finance},
  volume={66},
  number={1},
  pages={35--65},
  year={2011},
  publisher={Wiley},
  doi={10.1111/j.1540-6261.2010.01625.x}
}

@article{price2012earnings,
  title={Earnings Conference Calls and Stock Returns: The Incremental Informativeness of Textual Tone},
  author={Price, Steven M. and Doran, James S. and Peterson, David R. and Bliss, Barbara A.},
  journal={Journal of Banking \& Finance},
  volume={36},
  number={4},
  pages={992--1011},
  year={2012},
  publisher={Elsevier},
  doi={10.1016/j.jbankfin.2011.10.013}
}

@inproceedings{devlin2019bert,
  title={BERT: Pre-training of Deep Bidirectional Transformers for Language Understanding},
  author={Devlin, Jacob and Chang, Ming-Wei and Lee, Kenton and Toutanova, Kristina},
  booktitle={Proceedings of the 2019 Conference of the North American Chapter of the Association for Computational Linguistics},
  pages={4171--4186},
  year={2019}
}

@article{araci2019finbert,
  title={FinBERT: Financial Sentiment Analysis with Pre-trained Language Models},
  author={Araci, Dogu},
  journal={arXiv preprint arXiv:1908.10063},
  year={2019},
  url={https://arxiv.org/abs/1908.10063}
}

\clearpage
\section*{Appendices}
\subsection*{Individual Contributions}

    \begin{center}
    \begin{tabularx}{\textwidth}{|l|X|X|}
    \hline
    \textbf{Name} & \textbf{Role Area} & \textbf{Key Contributions} \\ \hline
    
    Owen Nyo Wei Yuan & Strategy \& Design, Report &
    Contributed to the overall project strategy and system design. Assisted in structuring the methodology and preparing sections of the final report. \\ \hline
    
    Ryan Tan Jun Wei & Optimisation, Exploratory Data Analysis (EDA), Report &
    Performed exploratory data analysis to understand dataset patterns and distributions. Implemented optimisation techniques to improve computational efficiency and contributed to report writing. \\ \hline
    
    Victor Tan Jia Xuan & Strategy \& Design, Report &
    Contributed to the formulation of the project strategy and design of the solution approach. Assisted in documenting the methodology, system design, and results in the final report. \\ \hline
    
    Fabian Ong Jun Yao & Data Cleaning, Exploratory Data Analysis (EDA), Report &
    Handled data preprocessing and cleaning to ensure dataset quality. Conducted exploratory data analysis and contributed to documentation in the final report. \\ \hline
    \end{tabularx}
    
    \captionof{table}{Individual Contributions of Team Members}
    \end{center}

\clearpage
\subsection*{Enhanced EDA Visualizations}
\renewcommand{\thefigure}{A\arabic{figure}}
\setcounter{figure}{0}

    \begin{figure}[H]
    \centering
    \includegraphics[width=0.7\textwidth]{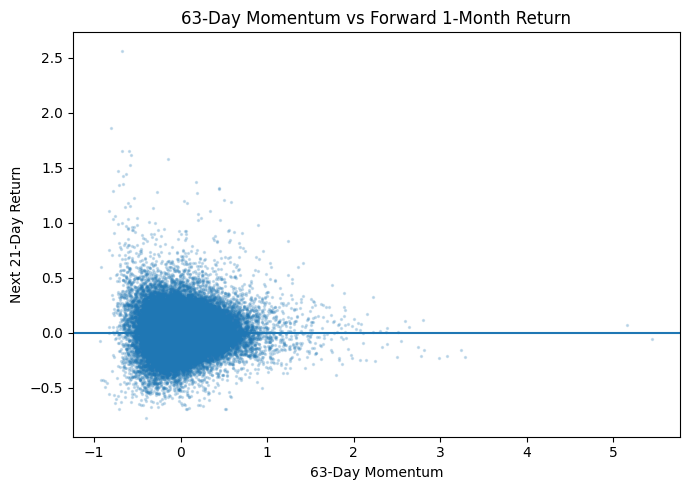}
    \caption{63-Day Momentum vs Forward 21-Day Return}
    \label{fig:momentum_vs_returns}
    \end{figure}
    
    \begin{figure}[H]
    \centering
    \includegraphics[width=0.7\textwidth]{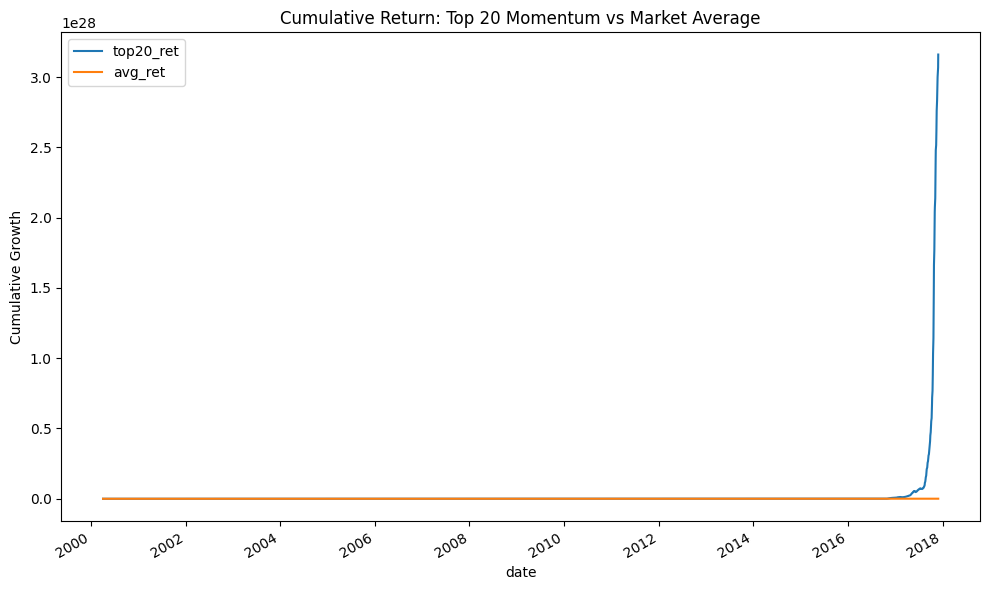}
    \caption{Cumulative Returns: Top Momentum Portfolio vs Market Average}
    \label{fig:cumulative_returns}
    \end{figure}
    
    \begin{figure}[H]
    \centering
    \includegraphics[width=0.7\textwidth]{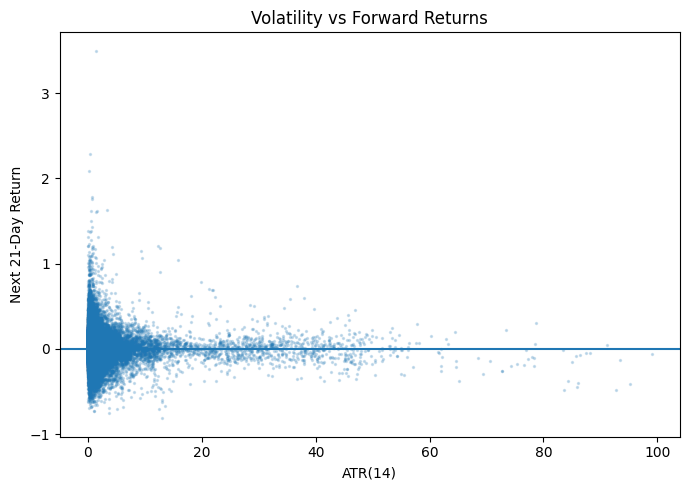}
    \caption{ATR(14) Volatility vs Forward 21-Day Return}
    \label{fig:atr_volatility}
    \end{figure}
    
    \begin{figure}[H]
    \centering
    \includegraphics[width=0.7\textwidth]{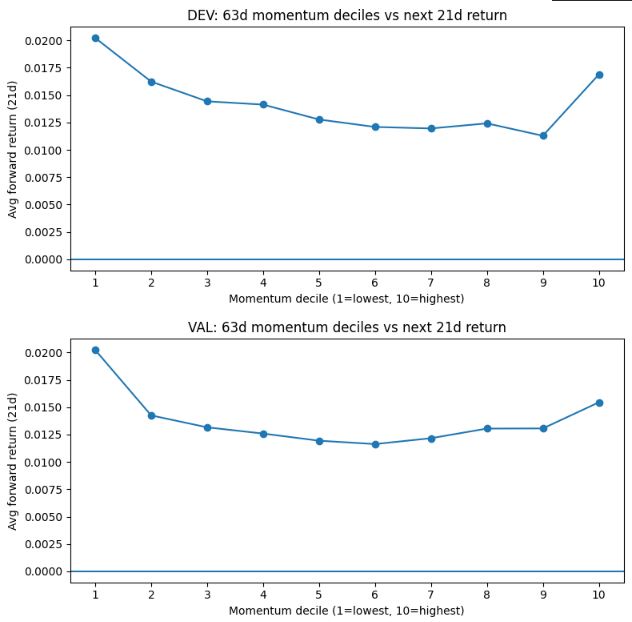}
    \caption{Momentum Deciles vs Average 21-Day Forward Return}
    \label{fig:momentum_decile}
    \end{figure}
    
    \begin{figure}[H]
    \centering
    \includegraphics[width=0.7\textwidth]{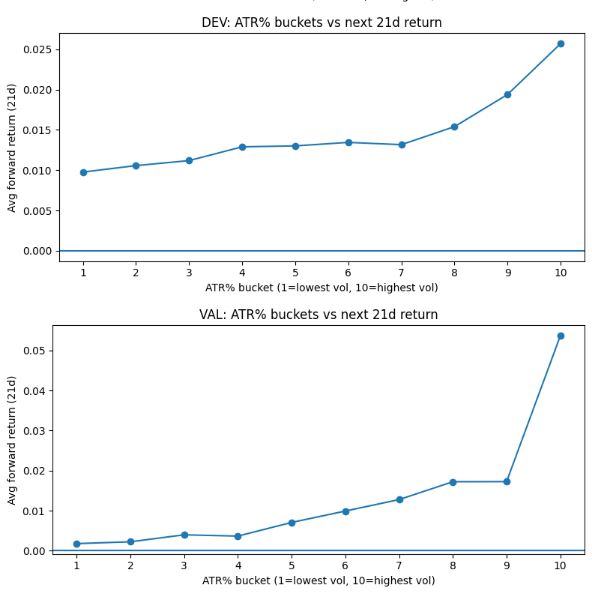}
    \caption{ATR Volatility Buckets vs Average 21-Day Forward Return}
    \label{fig:atr_bucket}
    \end{figure}
    
    \begin{figure}[H]
    \centering
    \includegraphics[width=0.7\textwidth]{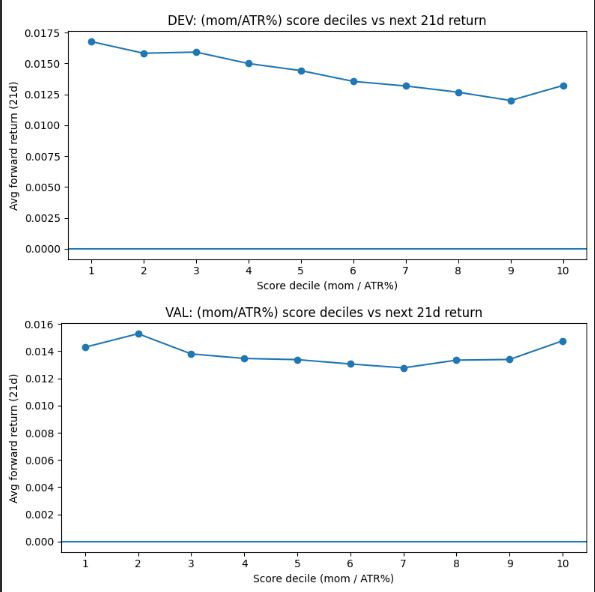}
    \caption{Momentum-to-Volatility Ratio Deciles vs Average 21-Day Forward Return}
    \label{fig:mom_atr_ratio}
    \end{figure}

\end{document}